\documentclass[letterpaper,journal]{IEEEtran}
\usepackage{amsmath,amsfonts}
\usepackage{algorithmic}
\usepackage{algorithm}
\usepackage{array}
\usepackage[caption=false,font=normalsize,labelfont=sf,textfont=sf]{subfig}
\usepackage{textcomp}
\usepackage{stfloats}
\usepackage{url}
\usepackage{verbatim}
\usepackage{graphicx}
\usepackage{cite}
\usepackage{multirow}
\usepackage{color}
\usepackage{makecell}
\usepackage{booktabs}
\usepackage{siunitx}
\usepackage[dvipsnames]{xcolor}
\begin{document}

\title{SeisMoLLM: Advancing Seismic Monitoring via Cross-modal Transfer with Pre-trained Large Language Model}

\author{Xinghao Wang, Feng Liu, Rui Su, Zhihui Wang, Lihua Fang, Lianqing Zhou, Lei Bai, Wanli Ouyang
        

\thanks{
Xinghao Wang is with the Shanghai Artificial Intelligence Laboratory, Shanghai 200232, China, and also with the International School of Information Science and Engineering, Dalian University of Technology, Dalian 116620, China (e-mail: wangxinghao@pjlab.org.cn).

Feng Liu is with the School of Electronic Information and Electrical Engineering, Shanghai Jiao Tong University, Shanghai 200240, China, and also with the Shanghai Artificial Intelligence Laboratory, Shanghai 200232, China (e-mail: liufeng2317@sjtu.edu.cn).

Rui Su, Lei Bai and Wanli Ouyang are with the Shanghai Artificial Intelligence Laboratory, Shanghai 200232, China (e-mail: surui@pjlab.org.cn; bailei@pjlab.org.cn; ouyangwanli@pjlab.org.cn).

Zhihui Wang is with the International School of Information Science and Engineering, Dalian University of Technology, Dalian 116620, China (e-mail: zhwang@dlut.edu.cn).

Lihua Fang and Lianqing Zhou are with the Institute of Earthquake Forecasting, China Earthquake Administration, Beijing 100036, China (e-mail: fanglihua@ief.ac.cn; zhoulq@ief.ac.cn).

Code is available at https://github.com/StarMoonWang/SeisMoLLM

}}



\maketitle

\begin{abstract}

    Recent advances in deep learning have revolutionized seismic monitoring, yet developing a foundation model that performs well across multiple complex tasks remains challenging, particularly when dealing with degraded signals or data scarcity. This work presents SeisMoLLM, the first foundation model that utilizes cross-modal transfer for seismic monitoring, to unleash the power of large-scale pre-training from a large language model without requiring direct pre-training on seismic datasets. Through elaborate waveform tokenization and fine-tuning of pre-trained GPT-2 model, SeisMoLLM achieves state-of-the-art performance on the DiTing and STEAD datasets across five critical tasks: back-azimuth estimation, epicentral distance estimation, magnitude estimation, phase picking, and first-motion polarity classification. It attains 36 best results out of 43 task metrics and 12 top scores out of 16 few-shot generalization metrics, with many relative improvements ranging from 10\% to 50\%. In addition to its superior performance, SeisMoLLM maintains efficiency comparable to or even better than lightweight models in both training and inference. These findings establish SeisMoLLM as a promising foundation model for practical seismic monitoring and highlight cross-modal transfer as an exciting new direction for earthquake studies, showcasing the potential of advanced deep learning techniques to propel seismology research forward.

\end{abstract}


\section{Introduction}

    \IEEEPARstart{S}{eismic} monitoring provides near-real-time analysis of earthquake impacts, playing an important role in supporting earthquake early warning systems to safeguard public safety. And it contributes to the creation of elaborate earthquake information catalogs, which are essential for advancing the study of seismic activities and improving our understanding of earthquake behaviors. 
    The rise of deep learning (DL)\cite{dl} techniques has revolutionized seismic monitoring in recent decades. DL-based approaches consistently outperform the conventional methods, achieving notable success in various seismological tasks\cite{general_review}, such as earthquake detection\cite{convnetquake}\cite{cred}, phase picking \cite{gpd, phasenet, eqt}, the estimations of earthquake magnitude\cite{magnet} and location parameters of epicentral distance and back-azimuth\cite{conv_parameterization}\cite{baznet}, and first-motion polarity classification\cite{p_and_polarity}\cite{japan_polarity}. While these methods have demonstrated success, they always encounter a set of inherent challenges linked to the vanilla supervised training approach, such as performance bottleneck in highly complex tasks like earthquake location, the difficulty in attaining robust performance in challenging scenarios, limited generalization with few training data, and the necessity of designing and training specialized neural networks from scratch for each monitoring task, which is notably time-consuming and resource-intensive.

    Recently, the strategy of pre-training and fine-tuning has shown exceptional performance across many domains in enhancing the performance and generalization capabilities of downstream tasks while lessening the dependence on target data. Transformer-based\cite{transformer} pre-trained models have achieved remarkable achievements in natural language processing\cite{bert, gpt2}, computer vision\cite{clip, sam, dall-e}, and other fields\cite{alphafold}. By processing data from various modalities into token sequences suitable for the Transformer architecture and conducting pre-training on vast amounts of data, these methods endow large models with excellent feature learning capabilities, strong few-shot generalization abilities, and the capacity to adapt to diverse downstream tasks through fine-tuning. These success motivate us to explore the potential of pre-training for tackling the fundamental challenges of seismic monitoring.
    
    However, seismic waveforms, as time series data, face the same challenge of in-domain heterogeneity that is prevalent in universal time-series pre-training \cite{ofa}\cite{calf}. Unlike language or images, which have established semantics and uniform formats, seismic waveforms exhibit significant heterogeneity due to variations in sampling methods, pre-processing techniques, noise levels, geological conditions, and labeling standards. These inconsistencies hinder the integration of existing datasets for large-scale pre-training on seismic waveforms, making it tricky to develop foundation models through direct pre-training and still in the early stages\cite{seislm}\cite{seisclip}.
    
    A promising alternative is to transfer the powerful sequence modeling capabilities of large Transformer-based models, originally pre-trained in other domains, to seismic monitoring. This cross-modal transfer strategy has demonstrated success in various fields\cite{voice2series, visionts, llm2code, llm2protein, dino2geo, cross-modal_finetune, multimodal}. And a series of recent works based on pre-trained large language model (LLM)\cite{gpt2}\cite{llama2} for time series tasks, further support this idea. Inspired by these developments\cite{ofa, test, time-llm, tempo, s2ip-llm, calf}, we choose GPT-2, the first LLM renowned for its remarkable performance achieved through large-scale pre-training, as our backbone. By freezing most of the parameters within its blocks and fine-tuning only a small subset, we successfully adapt its exceptional sequence feature extraction and few-shot generalization capabilities to seismic monitoring tasks.
    
    In this work, we propose SeisMoLLM, a novel seismic monitoring foundation model leveraging the idea of cross-modal transfer. This approach unlocks the potential of pre-training when doing so from scratch on seismic waveforms is infeasible. By leveraging the powerful feature learning and generalization capabilities acquired through pre-training on massive data from another domain, we address challenges that conventional network architecture modifications have struggled to overcome. These include accurately locating earthquakes with only single station, achieving reliable performance under poor data conditions, and generalizing to unseen data with few-shot training. The contributions of this work can be summarized as follows:
    
    1. \textbf{Pioneering Cross-Modal Transfer}: To the best of our knowledge, this is the first work to apply cross-modal transfer strategy in seismic monitoring, by fine-tuning a pre-trained large model, a novel approach is presented to mitigating longstanding challenges in the deep learning methods of seismic monitoring.
    
    2. \textbf{State-of-the-Art Performance}: Comparing with existing advanced methods on two widely used benchmark datasets, SeisMoLLM achieves state-of-the-art performance and few-shot generalization capabilities across multiple seismic monitoring tasks.
    
    3. \textbf{Efficiency and Practicality}: Despite employing a large pre-trained model, SeisMoLLM maintains inference efficiency comparable to baselines with much smaller models and achieves even better training efficiency, ensuring feasibility for real-world applications.
    
    4. \textbf{Adequate Experiments and Insights}: Through comprehensive experimental comparisons and ablation studies, we further demonstrate the promise of pre-trained LLM as a powerful feature extractors for seismic monitoring, and underscore the great potential of cross-modal transfer for future research.
    
    \begin{figure*}
        \centering
        \includegraphics[width=1.0\textwidth]{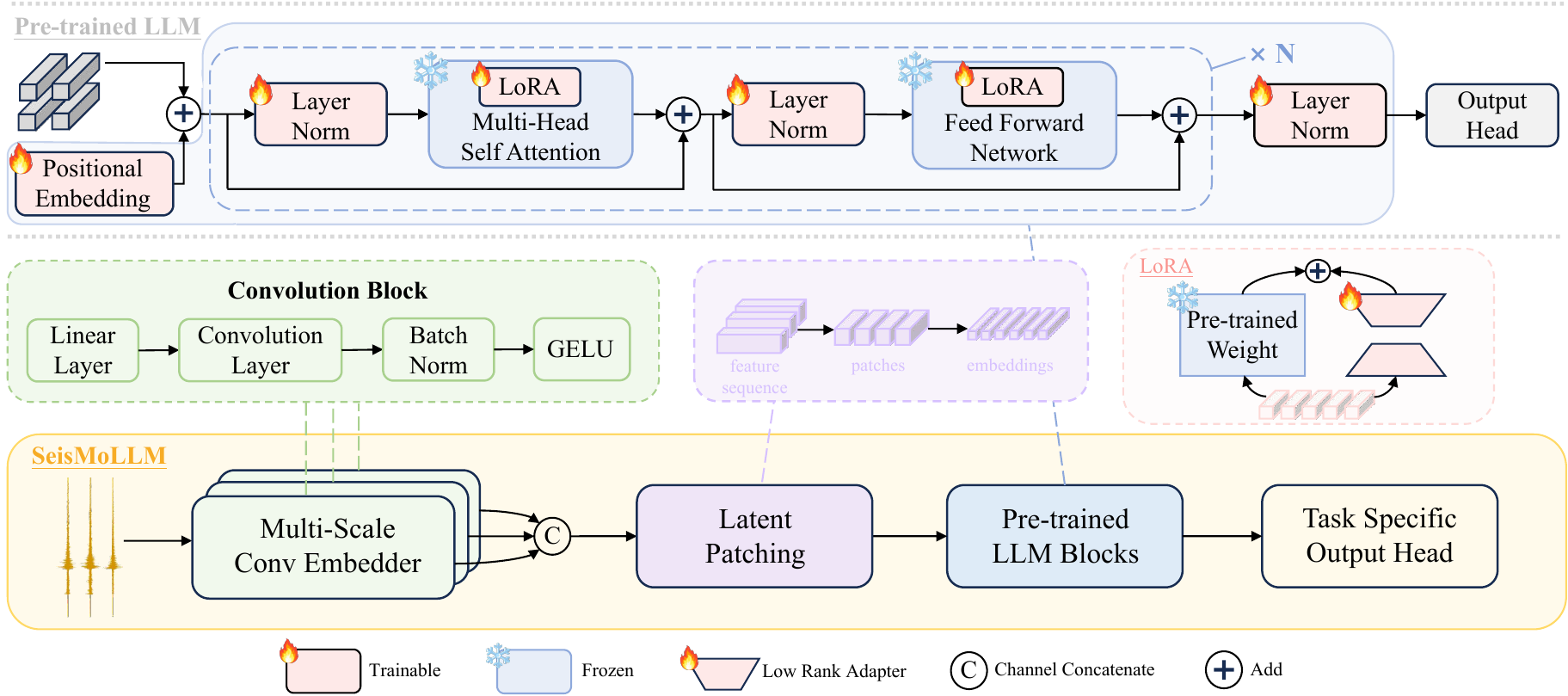}
        \caption{The architecture of proposed SeisMoLLM. Earthquake waveforms are processed by a multi-scale convolutional embedder to extract embeddings with local fine-grained features. These embeddings are then aggregated into much shorter token sequences via latent patching, which are fed into pre-trained LLM blocks with parameter-efficient fine-tuning for seismic monitoring. The corresponding output heads for different tasks generate the final results.}
        \label{model}
    \end{figure*}

\section{Methods}

    \subsection{Model Architecture}

        As illustrated in Figure \ref{model}, the architecture of SeisMoLLM is designed for seismic monitoring and consists of four key components: a multi-scale convolutional embedder, latent patching, pre-trained LLM blocks, and task-specific output heads. The embedder extracts multi-scale features from seismic waveforms, generating compact and informative embeddings. These embeddings are then aggregated into shorter token sequences through latent patching, which reduces the data volume while preserving critical information. The pre-trained LLM blocks receive the token sequence and employ their advanced sequence modeling capabilities for further feature learning. Finally, task-specific output heads generate the prediction results. The integration of these components enables SeisMoLLM to effectively process complex waveform data for accurate and efficient seismic monitoring. Detailed descriptions of each component are provided below. 
    
    \subsubsection{Multi-Scale Convolutional Embedder}
    
        The embedder is responsible for converting various input data into a form compatible with the network backbone. To leverage the powerful sequence modeling capabilities of LLM for seismic monitoring tasks, pre-trained Transformer blocks from LLM naturally serve as the primary feature extractor. Since the processing of waveform data before passing it through the LLM blocks is pivotal for optimal model performance \cite{first-step}, the embedder is result-driven designed. In this study, several approaches for embedding the waveform data were explored. One approach involved directly patching and linearly embedding the data. However, this method proved inadequate, as it struggled to balance the capture of local details with computational efficiency. Specifically, using larger patch sizes hindered the model's ability to learn fine-grained features, while excessively reducing patch sizes resulted in an impractically large number of tokens, incurring significant computational costs. Another approach leverages convolutions, which inherently specialize in learning local fine-grained features and compressing sequence length. This convolution-based method not only complements the global feature learning capabilities of the LLM blocks but also addresses the need to reduce the number of subsequent tokens, thereby offering a more efficient solution.
        
        Considering that different tasks focus on varying local scales, we designed a multi-scale convolution block, it can be formally written as:
        \begin{equation}
            \boldsymbol{Y} = \text{BN}(\text{Proj}(\text{Concat}([\boldsymbol{X}_{0}, \boldsymbol{X}_{1}, \dots, \boldsymbol{X}_{n}]))),
        \end{equation}
        where \( \boldsymbol{X}_{0}, \boldsymbol{X}_{1}, \dots, \boldsymbol{X}_{n} \) represent the feature sequences obtained from $n$ convolutional modules with various kernel sizes, and this module is defined as:
        \begin{equation}\begin{aligned}
            \boldsymbol{X}_{i} &= \text{GELU}(\text{BN}(\text{Conv}_{k_i}(\text{Proj}(\boldsymbol{X})))), \\
            \forall i &\in \{0, 1, \dots, n\}.
        \end{aligned}\end{equation}
        In the above equations, Proj represents a linear transformation implemented by a linear layer, $\text{Conv}_{k_i}$ denotes a convolution layer with a kernel size of \(k_i\), BN refers to a BatchNorm\cite{batchnorm} layer, GELU\cite{gelu} is the activation function, and Concat indicates concatenation along the channel dimension.
        
        By stacking these blocks before LLM as an embedder, SeisMoLLM can reduce the feature sequence length while effectively learning multi-scale local details with minimal computation, ensuring the feasibility and effectiveness of following LLM blocks.
    
    \subsubsection{Latent Patching}
        
        Patching is one of the most effective and simple methods to aggregate local information in Transformer-based methods \cite{patchtst}\cite{vit}, where both time and space complexity are proportional to the square of the number of tokens.
        However, applying fixed patching directly to the original input data before the embedder hinders the learning of multi-scale local features. Furthermore, compared to first applying convolutions to learn from the continuous seismic waveform, the discrete partitioning operation of patching can potentially lose some local details of the data \cite{hdmixer}.
        
        Therefore, we applied latent patching with a patch size and stride of 8 to the feature sequence before LLM Blocks. Starting with the feature sequence \( \boldsymbol{x} \in \mathbb{R}^{C \times L} \), where \(C\) is the number of channels and \(L\) is the sequence length, latent patching first reshapes it to \( \boldsymbol{x} \in \mathbb{R}^{C \times N \times P} \), where \(N = \frac{T}{P} \) represents the number of patches, each with a size of \(P\). Subsequently, features from \(P\) consecutive points in the sequence are aggregated, resulting in the final token sequence \( \boldsymbol{x} \in \mathbb{R}^{N \times (C \times P)} \). This also reduces the number of tokens fed into the LLM blocks by a factor of \(P\), significantly lowering the computational load. 

    \subsubsection{Pre-trained LLM Blocks}
        Large language models are very large deep learning models that are pre-trained with self-supervised learning on vast amounts of data for natural language processing tasks. Studies have shown that pre-training on extensive text corpora equips these models with a general capability for token sequence modeling, which can be effectively transferred to other modalities\cite{llm2code, llm2protein, ofa}. Therefore, pre-trained LLM blocks serve as the core component and the structural backbone in our method for seismic monitoring tasks. Specifically, we conservatively choose the smallest version of GPT-2 model with 12 layers and hidden dimension of 768\cite{gpt2} as our model backbone instead of more advanced LLMs to demonstrate that even the earliest LLM holds enormous potential for transferring to seismic domain, and no need for special model designs or extremely strong performance. We retain the complete structure of the decoder blocks in GPT-2 and freeze most of the pre-trained parameters to preserve its capabilities and generalization ability during cross-modal transfer.
        
        To transform the LLM blocks from text token processor to seismogram patch embedding learner, we set the positional embeddings as trainable to fine-tune the LLM blocks' understanding of the positional relationships within seismic waveform data. We also fine-tune the layer normalization layers to adapt to the feature distributions of new modality. Since the self-attention layers and feed-forward networks (FFN) form the core of feature learning within LLM blocks and contain the majority of the generalized representation learning capabilities acquired through pre-training, we freeze the pre-trained parameters of them, and employ Low-Rank Adaptation (LoRA)\cite{lora} as insert modules to further fine-tune these two components to efficiently capture seismic waveform features.
        
        LoRA introduces low-rank updates to the weight matrices of large pre-trained models, instead of directly updating all the parameters during fine-tuning, thus significantly reducing the number of trainable parameters. LoRA can be formulated as: 
        \begin{equation}
            \boldsymbol{W}_{\text{adapted}} = \boldsymbol{W} + \Delta\boldsymbol{W}, \Delta\boldsymbol{W} = \boldsymbol{A}\boldsymbol{B},
        \end{equation} where \(\boldsymbol{W} \in \mathbb{R}^{d \times k}\) is the original weight matrix, and \(\Delta\boldsymbol{W} \in \mathbb{R}^{d \times k}\) represents the low-rank update. The low-rank decomposition consists of two learnable matrices: \(\boldsymbol{A} \in \mathbb{R}^{d \times r}\) and \(\boldsymbol{B} \in \mathbb{R}^{r \times k}\), where their rank $r \ll \text{min}(d, k)$. 
        By constraining \(\Delta\boldsymbol{W}\) to a low-rank form, LoRA enables domain adaptation with minimal additional parameters, thereby lowering memory and computational requirements, while also maintaining performance comparable to full fine-tuning. And based on all above, SeisMoLLM can achieve cross-modal parameter-efficient fine-tuning by setting only 10\% of its parameters as trainable.

    \subsubsection{Task-Specific Heads}
        
        The output head functions to decoding and summarizing the feature sequence at the end of the neural network to generate the final task prediction. For task-specific output heads, we adopted classic and simple structures, similar to the design of \cite{seist}.
        In the phase picking task, upsampling and convolutional blocks are used to progressively restore the latent feature sequence to the original data shape. A sigmoid function is then applied to generate two probability sequences, corresponding to the arrival of P and S phases.
        
        For other regression and classification tasks, stacked convolutional layers form a fully convolutional block to further aggregate information from the feature sequence. The output is then obtained through global average pooling, flattening, linear projection, and the corresponding activation functions. Specifically, SeisMoLLM predicts the sine and cosine values of back-azimuth as learning targets, using the tanh activation function to constrain the output to (-1, 1). Softmax function is applied to generate 2D one-hot encodings representing first-motion polarity (upward or downward). For magnitude and epicentral distance estimation, sigmoid functions provide outputs in the range (0, 1), which are then scaled to obtain the final predictions.

    \subsection{Dataset and Label}
    
        To enable intuitive and fair comparisons with existing methods and to comprehensively demonstrate the versatility of our method across diverse geological scenarios and data conditions, we selected two datasets with distinct geographic distributions and pre-processing methods. The first is the STanford EArthquake Data Set (STEAD), which comprises 1030k three component earthquake waveforms of global seismic events with a sampling rate of 100 Hz\cite{stead}. The second is the DiTing dataset, which comprises 2707k earthquake waveforms of 50 Hz sampling from China and surrounding regions, including large amounts of data with low signal-to-noise ratios\cite{diting}. Both datasets were utilized for model training and testing.
    
        \begin{table}[!h]
        \caption{Dataset Splits and Details\label{data}}
        \centering
        \renewcommand\arraystretch{2}
        \setlength{\tabcolsep}{1.2mm}{
        \begin{tabular}{lcccccc}
        \hline
        \specialrule{0pt}{-2pt}{-2pt}
        Dataset & Setting & Task & All & Train & Val & Eval \\
        \hline
        \multirow{4}{*}{DiTing} 
        & Standard & All tasks & 277737 & \makecell{222189 \\ (80\%)} & \makecell{27773 \\ (10\%)} & \makecell{27775 \\ (10\%)} \\ 
        \cline{2-7}
        \specialrule{0pt}{0pt}{0.6pt}
        & \multirow{3}{*}{Few-shot} 
        & \makecell{picking \& \\ magnitude} & 2706934 & \makecell{222189 \\ (8\%)} & \makecell{27773 \\ (1\%)} & \makecell{2456972 \\ (91\%)} \\ 
        \cline{3-7}
        \specialrule{0pt}{0pt}{0.6pt}
        & & \makecell{back-azimuth \\ \& distance} & 2662729 & \makecell{222189 \\ (8\%)} & \makecell{27773 \\ (1\%)} & \makecell{2412767 \\ (91\%)} \\
        \cline{3-7}
        & & polarity & 613244 & {\makecell{222189 \\ (36\%)}} & {\makecell{27773 \\ (5\%)}} & {\makecell{363282 \\ (59\%)}} \\
        \hline
        \multirow{4}{*}{STEAD} 
        & \multirow{2}{*}{Standard} & magnitude & 725298 & {\makecell{580238 \\ (80\%)}} & {\makecell{72529 \\ (10\%)}} &{\makecell{72531 \\ (10\%)}} \\ 
        \cline{3-7}
        & & other 6 & 1030232 & {\makecell{824185 \\ (80\%)}} & {\makecell{103023 \\ (10\%)}} & {\makecell{103024 \\ (10\%)}} \\ 
        \cline{2-7}
        & \multirow{2}{*}{Few-shot} 
        & magnitude & 725298 & {\makecell{72530 \\ (10\%)}} & {\makecell{36264 \\ (5\%)}} &{\makecell{616504 \\ (85\%)}} \\ 
        \cline{3-7}
        & & other 6 & 1030232 & {\makecell{103024 \\ (10\%)}} & {\makecell{51511 \\ (5\%)}} &{\makecell{875697 \\ (85\%)}} \\ 
        \hline
        \end{tabular}}
        \end{table}

        Notably, due to the varied completeness of labels for different tasks in DiTing dataset, we followed \cite{seist} and used the subset with complete labels for all tasks in this work, which has 278k waveform samples. The remaining data was used for few-shot generalization testing. And due to the differences in label types, we only selected data with local magnitude (ML) labels for magnitude estimation in STEAD dataset. More dataset splits and details for each task are provided in Table \ref{data}.
        
        For labels of each task, we used probability sequences indicating the likelihood of phase arrivals for phase picking. To ensure a fair comparison with baselines, Gaussian-shaped labels were employed: the probabilities for the arrival times of P-wave and S-wave were set to 1 at their respective labeled positions and gradually decreased to 0 before and after these point, adhering to a Gaussian distribution with a total width of 0.5 seconds. For first-motion-polarity, 2D one-hot encodings were used to indicate upward or downward. And all three regression tasks took ground truth values as their labels. 

    \subsection{Training and Evaluation}
    
        In the training process, common data augmentation techniques were applied to waveform data, including adding random Gaussian noise, time drift, introducing gaps, channel dropout, amplitude scaling, pre-emphasis, and noise generation, with probabilities of 0.4, 0.4, 0.4, 0.4, 0.4, 0.97, and 0.05, respectively, following the baseline method\cite{seist}. The LoRA modules were configured with biases, both rank and alpha are set to 16, and dropout with a probability of 0.1 is applied to both LLM blocks and LoRA.
    
        For picking P and S phases, we used the sum of the respective binary cross-entropy loss to train the whole task, as formulated in 
        \begin{equation}
            \mathcal{L}_{\text{BCE}} = -\frac{1}{N} \sum_{i=1}^N \big( y_{i} \log(\hat{y}_{i}) + (1 - y_{i}) \log(1 - \hat{y}_{i}) \big),
        \end{equation}
        \vspace{-1.6mm}
        \begin{equation}
            \mathcal{L}_{\text{picking}} = \mathcal{L}_{\text{BCE}}^{\text{P}} + \mathcal{L}_{\text{BCE}}^{\text{S}}.
        \end{equation}
        Specifically, for a set of \(N\) samples, \(y_{i} \in [0, 1]\) represents the ground truth label, and \(\hat{y}_{i} \in [0, 1]\) is the predicted probability. For the phase picking task loss, \(\mathcal{L}_{\text{BCE}}^{\text{P}}\), \(\mathcal{L}_{\text{BCE}}^{\text{S}}\) denote the BCE losses for P and S phases picking, respectively.
        Cross-entropy loss as shown in 
        \begin{equation}
            \mathcal{L}_{\text{CE}} = -\frac{1}{N} \sum_{i=1}^N y_{i} \log(\hat{y}_{i})
        \end{equation} 
        was used for the first-motion polarity classification task. Similarly, \(y_{i}\) and \(\hat{y}_{i}\) maintain their definitions, representing the binary label and predicted probability, respectively. 
        For regression tasks, Huber loss as shown in 
        \begin{equation}
            \mathcal{L}_{\text{Huber}} = \frac{1}{N} \sum_{i=1}^N 
            \begin{cases} 
                \text{0.5} (y_{i} - \hat{y}_{i})^2, & \text{if } |y_{i} - \hat{y}_{i}| \leq \delta, \\
                \delta |y_{i} - \hat{y}_{i}| - \text{0.5} \delta, & \text{if } |y_{i} - \hat{y}_{i}| > \delta,
            \end{cases}
        \end{equation} 
        was adopted, where \(y_{i} \in \mathbb{R}\) is the true value, \(\hat{y}_{i} \in \mathbb{R}\) is the predicted value, and \(\delta\) is set to 1 here.
        Training employed the Adam optimizer\cite{adam} and cyclic learning rate scheduler\cite{cycliclr}, with the learning rate oscillating between 5×10\textsuperscript{-4} and 1×10\textsuperscript{-3} to avoid local optima. Early stopping was triggered if validation loss did not decrease for 30 consecutive epochs.
    
        Classification tasks were evaluated using Precision (Pr), Recall (Re), and F1 score (F1), while regression tasks were assessed using mean absolute error (MAE), coefficient of determination (R\textsuperscript{2}), mean error (Mean), and standard deviation (Std). Specifically, for phase picking, predictions with arrival errors  \textless0.1 seconds were considered positive samples for calculating classification metrics, and time point-wise errors were used for regression metrics. All metrics followed standard definitions, formula shown in Equations \ref{cls_metrics}, \ref{reg_metrics} in Appendix A.
    
        All experiments were conducted on an Ubuntu server equipped with NVIDIA RTX-4090 GPUs and AMD EPYC 7402 24-Core CPUs. Training experiments utilized 4 GPUs and 24 CPU cores, while evaluation experiments were performed using 1 GPU and 6 CPU cores.
    
\section{Results}

    \subsection{Task Performances}
    
\begin{figure*}
        \centering
        \includegraphics[width=1\textwidth]{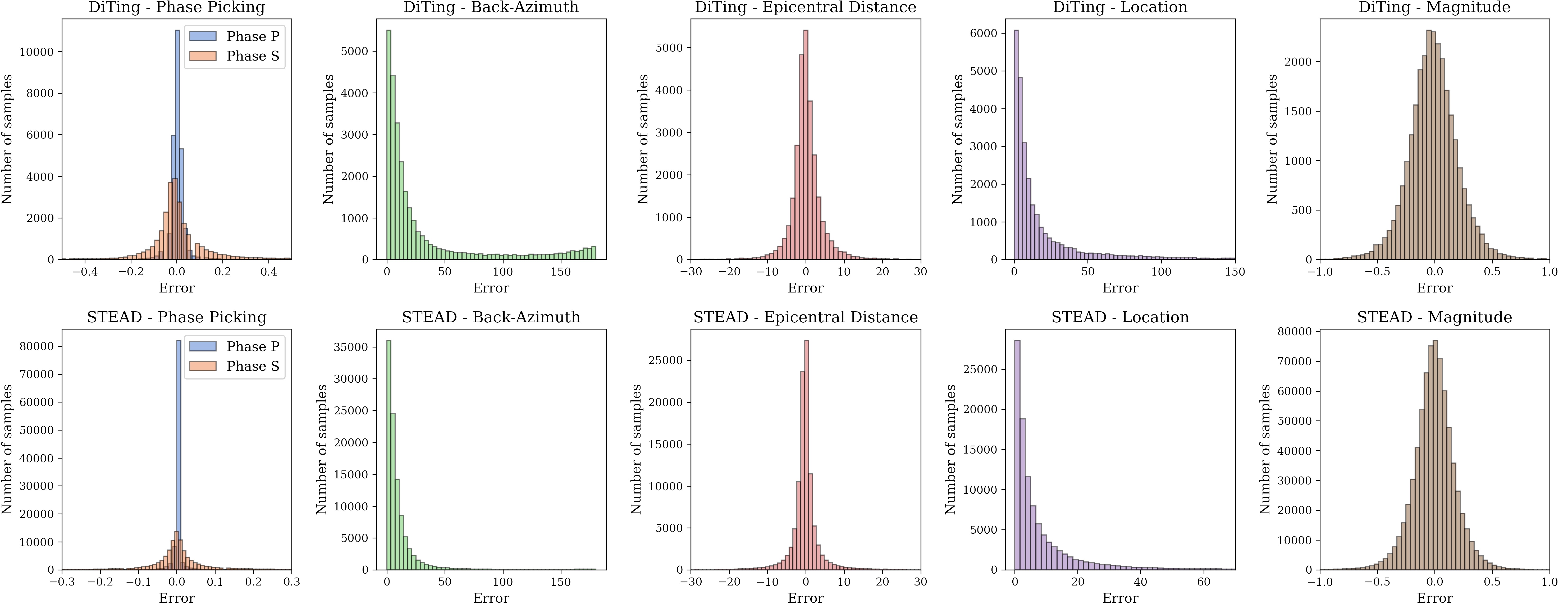}
        \caption{The residuals for phase picking are measured in seconds, back-azimuth estimation residuals are in degrees, and the units for epicentral distance and earthquake location residuals are kilometers. The first and second rows correspond to the results on the DiTing and STEAD datasets, respectively.} 
        \label{error_distribution}
        \end{figure*}

        In six key seismic monitoring tasks, including back-azimuth estimation, epicentral distance estimation, magnitude estimation, phase picking and first-motion polarity classification, SeisMoLLM achieved state-of-the-art performance on the majority of tasks across both the STEAD and DiTing datasets. Compared to advanced baseline methods such as the large-scale SeisT model (SeisT-L) \cite{seist}, BAZ network \cite{baznet}, MagNet \cite{magnet}, EQTransformer \cite{eqt}, and PhaseNet \cite{phasenet}, SeisMoLLM consistently demonstrated superior accuracy and robustness. Detailed results are presented in Tables \ref{performance1}\, \ref{performance2}, and Figure \ref{error_distribution} illustrates the residual distribution for all tasks except classification.
    
        \begin{table*}[!t]\small
        \begin{minipage}[t]{0.48\linewidth}
            \caption{Model Performance Comparison on STEAD Dataset \\ \textbf{\textcolor{red}{red}}: best, \textcolor{blue}{blue}: second best\label{performance1}}
            \centering
            \setlength{\extrarowheight}{1pt}
            \setlength{\tabcolsep}{1.74mm}
            \begin{tabular}{llcccc}
                \hline
                Task & Model & MAE & R\textsuperscript{2} & Mean & Std \\
                \hline    
                \multirow{3}{*}{\makecell{Back \\ Azimuth}}
                & BAZ network  & 44.489 & 0.614  & \textcolor{blue}{-0.384} & 63.491 \\
                & SeisT-L      & \textcolor{blue}{23.890} & \textcolor{blue}{0.824} & -0.601 & \textcolor{blue}{42.888} \\
                & SeisMoLLM    & \textcolor{red}{11.342} & \textcolor{red}{0.948} & \textcolor{red}{-0.076} & \textcolor{red}{23.390} \\ 
                \hline
                \multirow{2}*{Distance} 
                & SeisT-L      & \textcolor{blue}{2.609} & \textcolor{blue}{0.988} & \textcolor{blue}{-0.175} & \textcolor{blue}{5.413} \\
                & SeisMoLLM    & \textcolor{red}{2.313} & \textcolor{red}{0.990} & \textcolor{red}{0.169} & \textcolor{red}{4.843} \\
                \hline
                \multirow{3}*{Magnitude}
                & MagNet      & 0.225 & 0.897 & \textcolor{red}{-0.003} & 0.313 \\ 
                & SeisT-L     & \textcolor{blue}{0.172} & \textcolor{red}{0.939} & 0.016 & \textcolor{blue}{0.242} \\
                & SeisMoLLM   & \textcolor{red}{0.158} & \textcolor{blue}{0.931} & \textcolor{blue}{0.007} & \textcolor{red}{0.221} \\
                \hline
                Task & Model & F1 & Pr & Re & MAE \\
                \hline
                \multirow{4}*{Phase P}
                & PhaseNet          & 97.79 & 97.82 & 97.75 & 0.556 \\
                & EQTransformer     & \textcolor{blue}{98.14} & 98.14 & \textcolor{blue}{98.13} & 0.590 \\
                & SeisT-L           & \textcolor{red}{98.24} & \textcolor{red}{98.25} & \textcolor{red}{98.24} & \textcolor{blue}{0.555} \\
                & SeisMoLLM         & 98.08 & \textcolor{blue}{98.15} & 98.01 & \textcolor{red}{0.498} \\
                \hline
                \multirow{4}*{Phase S} 
                & PhaseNet          & 80.04 & 80.67 & 79.42 & \textcolor{blue}{2.296} \\
                & EQTransformer     & 80.57 & 81.14 & 80.00 & 2.400 \\
                & SeisT-L           & \textcolor{blue}{80.91} & \textcolor{blue}{81.40} & \textcolor{blue}{80.43} & 2.312 \\
                & SeisMoLLM        & \textcolor{red}{81.66} & \textcolor{red}{82.26} & \textcolor{red}{81.08} & \textcolor{red}{2.258} \\
                \hline
            \end{tabular}
            \vfill
        \end{minipage}
        \hspace{\fill}
        \begin{minipage}[t]{0.48\linewidth}
            \caption{Model Performance on DiTing Dataset \\ \textbf{\textcolor{red}{red}}: best, \textcolor{blue}{blue}: second best\label{performance2}}
            \centering
            \setlength{\extrarowheight}{1pt}
            \setlength{\tabcolsep}{1.74mm}
            \begin{tabular}{llcccc}
                \hline
                Task & Model & MAE & R\textsuperscript{2} & Mean & Std \\
                \hline
                \multirow{3}*{\makecell{Back \\ Azimuth}}
                & BAZ network      & 46.484 & 0.522 & \textcolor{blue}{0.253} & 71.410 \\
                & SeisT-L      & \textcolor{blue}{42.479} & \textcolor{blue}{0.565} & \textcolor{red}{0.108} & \textcolor{blue}{68.069} \\
                & SeisMoLLM & \textcolor{red}{33.551} & \textcolor{red}{0.676} & -0.533 & \textcolor{red}{58.734} \\
                \hline
                \multirow{2}*{Distance} 
                & SeisT-L      & \textcolor{blue}{3.126} & \textcolor{blue}{0.983} & \textcolor{blue}{-0.146} & \textcolor{blue}{7.386} \\
                & SeisMoLLM & \textcolor{red}{2.972} & \textcolor{red}{0.986} & \textcolor{red}{-0.139} & \textcolor{red}{6.829} \\
                \hline
                \multirow{3}*{Magnitude}
                & MagNet      & 0.193 & 0.927 & 0.025 & 0.266 \\
                & SeisT-L      & \textcolor{blue}{0.176} & \textcolor{blue}{0.940} & \textcolor{red}{-0.007} & \textcolor{blue}{0.240} \\
                & SeisMoLLM & \textcolor{red}{0.166} & \textcolor{red}{0.947} & \textcolor{blue}{-0.012} & \textcolor{red}{0.227} \\
                \hline
                Task & Model & F1 & Pr & Re & MAE \\
                \hline
                \multirow{4}*{Phase P} 
                & PhaseNet           & 94.75 & \textcolor{blue}{95.48} & 94.03 & \textcolor{blue}{0.913} \\
                & EQTransformer      & 94.57 & 95.03 & 94.12 & 1.053 \\
                & SeisT-L            & \textcolor{blue}{94.99} & 95.38 & \textcolor{blue}{94.61} & 0.990 \\
                & SeisMoLLM          & \textcolor{red}{95.69} & \textcolor{red}{96.16} & \textcolor{red}{95.24} & \textcolor{red}{0.785} \\
                \hline
                \multirow{4}*{Phase S} 
                & PhaseNet           & 69.02 & 70.88 & 67.26 & \textcolor{blue}{1.312} \\
                & EQTransformer      & 68.57 & 70.14 & 67.06 & 1.339 \\
                & SeisT-L            & \textcolor{blue}{69.97} & \textcolor{blue}{71.27} & \textcolor{blue}{68.72} & 1.329 \\
                & SeisMoLLM          & \textcolor{red}{72.82} & \textcolor{red}{73.66} & \textcolor{red}{72.01} & \textcolor{red}{1.277} \\
                \hline
                \multirow{3}*{\makecell{First \\ Motion}}
                & DiTingMotion       & 86.22 & 81.74 & 91.23 & - \\
                & SeisT-L            & \textcolor{blue}{93.81} & \textcolor{blue}{93.89} & \textcolor{blue}{93.74} & - \\
                & SeisMoLLM          & \textcolor{red}{94.33} & \textcolor{red}{94.42} & \textcolor{red}{94.25} & - \\
                \hline
            \end{tabular}
        \end{minipage}
        \vspace{5pt} 
        \par\noindent\makebox[\linewidth][c]{%
        \parbox{\linewidth}{
        All metrics are calculated using the same data and settings for training and evaluation. F1, Pr and Re are F1-score, Precision and Recall expressed as percentages. Mean and Std are the mean and standard deviation of errors. The error tolerance is 0.1 second, and MAE is based on the error in units of data points in the P and S phase picking tasks. All tasks use seismic waveforms of the original length as inputs.
        }}
        \end{table*}
    
        \begin{table}[!t]
        \caption{Single Station Localization Performance Comparison\label{location}}
        \centering
        \setlength{\extrarowheight}{1pt}
        \setlength{\tabcolsep}{1.8mm}
        \begin{tabular}{clcccc}
        \hline
        Dataset & Model & MAE & MAPE & RMSE & Std \\
        \hline
        \multirow{2}*{STEAD}
        & SeisT-L   & 21.593 & 0.437 & 49.644 & 44.702 \\
        & SeisMoLLM & 10.804 & 0.235 & 27.882 & 25.704 \\
        \hline
        \multirow{2}*{DiTing}
        & SeisT-L   & 38.026 & 0.682 & 81.087 & 71.619 \\
        & SeisMoLLM & 30.216 & 0.550 & 70.341 & 63.522 \\
        \hline
        \end{tabular} \\
        \vspace{5pt}
        \par\noindent\makebox[\linewidth][c]{%
        \parbox{\linewidth}{
        All metrics are calculated using results in km. MAPE, RMSE are mean absolute percentage error and root mean squared error. And the percentage error for MAPE is calculated as location error / epicentral distance.
        }}
        \end{table}

        \begin{figure*}[!ht]
        \centering
        \includegraphics[width=0.97\textwidth]{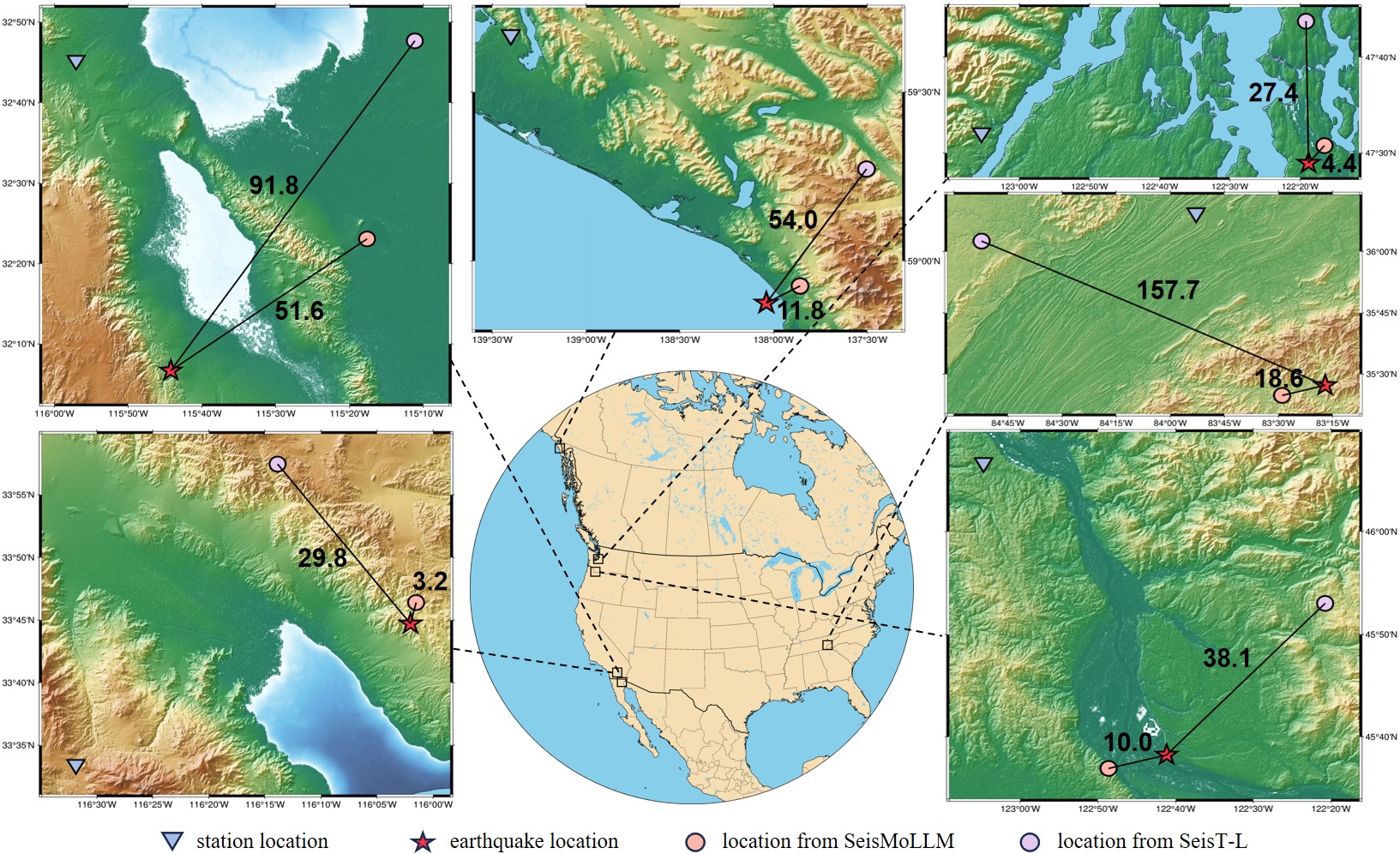}
        \caption{Comparison of earthquake location examples between SeisMoLLM and SeisT-L using events in the USA from STEAD dataset. The black line and the number above indicates the location error, measured in kilometers.} 
        \label{NA_location}
        \vspace{-1.5mm}
        \end{figure*}
    
        Especially for regression tasks related to earthquake location and magnitude, SeisMoLLM significantly outperformed baselines that were meticulously designed and trained from scratch for each individual task.
    
        For example, in the most challenging task of back-azimuth estimation, SeisMoLLM achieved an impressive mean absolute error (MAE) of 11.342° and coefficient of determination (R\textsuperscript{2}) of 0.948 on STEAD dataset, improving the best previous records by 53\% and 15\%, respectively, reaching a new level of performance. Similarly, substantial improvements of 21\% in MAE and 20\% in R\textsuperscript{2} were also observed on DiTing dataset.
    
        For epicentral distance estimation, the R\textsuperscript{2} of SeisMoLLM reached approximately 0.99 on both datasets, with MAE reduced to 2.313 km on STEAD and 2.972 km on DiTing, corresponding to reductions of 11\% and 5\%.
    
\begin{figure*}
        \centering
        \includegraphics[width=0.96\textwidth]{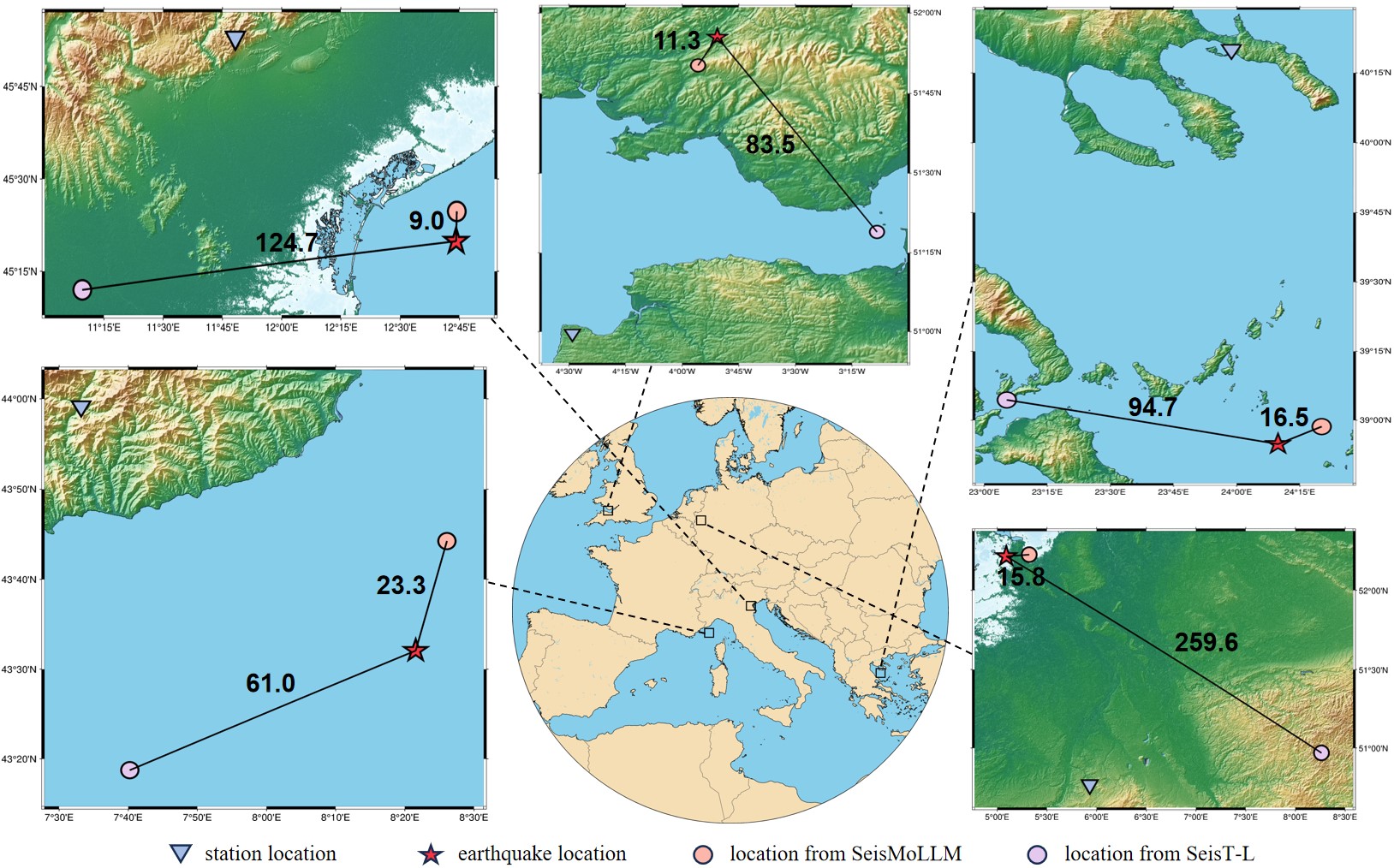}
        \caption{Comparison of earthquake location examples between SeisMoLLM and SeisT-L using events in Europe from STEAD dataset. The black line and the number above indicates the location error, measured in kilometers.} 
        \label{EU_location}
        \end{figure*}

        Furthermore, by combining the above two tasks, we can fairly compare and intuitively visualize the performance of SeisMoLLM and existing methods on single-station earthquake location, as shown in Table \ref{location}. Following the distribution of actual event locations within the 2 datasets, Figures \ref{NA_location} and \ref{EU_location} illustrate the location examples in STEAD dataset that are located in the USA and Europe, respectively. 
        On the STEAD dataset, SeisMoLLM model exhibits remarkable performance, with an MAE that is nearly half of SeisT's and a MAPE as low as 0.235, achieving significant improvements ranging from 43\% to 50\% across all metrics. For DiTing, SeisMoLLM further demonstrates its superiority, reducing both MAE and MAPE by approximately 20\% compared to the baseline. Visualization of examples from the USA and Europe underscore that our model consistently delivers more accurate predictions than SeisT. These results showcase the model's exceptional ability on the complex task of earthquake location, proving its strong capability to advance challenging tasks.
    
        Magnitude estimation further showcases the superior performance of our model on regression tasks, achieving an MAE of around 0.16 on both datasets, reducing error by 8\% and 7\% compared to the previous best models.
    
        Beyond regression tasks, SeisMoLLM also excelled in the classical phase-picking task. With the error tolerance of 0.1 second, it achieved the highest F1 scores of 0.957 and 0.728 for P and S phase picking on DiTing dataset, and also surpassing SeisT, delivering competitive performance on STEAD dataset. 
    
        The first-motion polarity classification task was conducted only on DiTing, due to the lack of labels in STEAD, and SeisMoLLM also achieved a state-of-the-art F1 score of 0.943.
    
        Notably, while previous methods require segmenting the seismogram around the P-phase to highlight relevant information for back-azimuth estimation and first-motion polarity classification \cite{baznet}\cite{seist}\cite{ditingmotion}, SeisMoLLM achieves the aforementioned superior performance directly from the input waveform without relying prior knowledge of P-wave arrival times, demonstrating the strong feature learning capability of our model.

    \subsection{Few-shot Generalization}
    
\begin{figure*}
        \centering
        \includegraphics[width=1\textwidth]{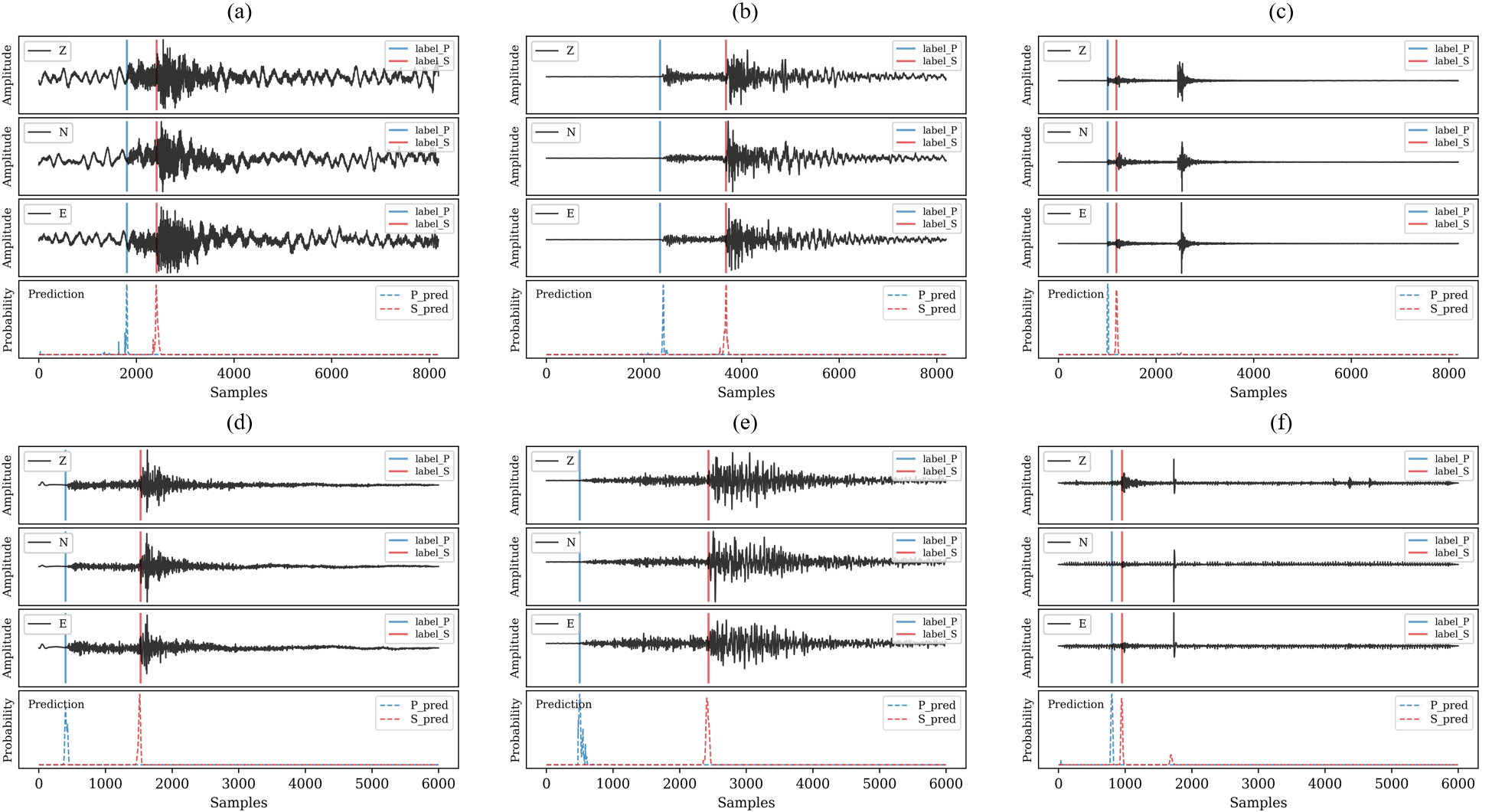}
        \caption{Few-shot phase picking examples. The three columns are representative of difficult samples with low signal-to-noise ratios (SNR), far epicentral distances, and low magnitudes, respectively. (a)-(c) are from DiTing dataset and (d)-(f) are from STEAD dataset. Figures are all drawn from the raw model output without any modifications.}
        \label{few-picking}
        \end{figure*}

        To demonstrate that our method effectively transfers the powerful sequence modeling capabilities of large language models from the natural language to seismic waveform, rather than merely fitting specific training data, we compared the generalization ability of SeisMoLLM with advanced baselines under a 10\% few-shot setting. The first-motion polarity classification task was not included as lacking sufficient data, and the results are shown in Table \ref{few-shot}, and Figure \ref{few-picking} presents examples of phase picking results from models trained with few-shot data.
    
        \begin{table}[!ht]\small
        \caption{10\% Few-shot Performance Comparison on Two Datasets \\ \textbf{\textcolor{red}{red}}: best, \textcolor{blue}{blue}: second best\label{few-shot}}
        \centering
        \setlength{\extrarowheight}{1pt}
        \setlength{\tabcolsep}{1.8mm}
        \begin{tabular}{llcccc}
            \hline
            \multicolumn{2}{c}{Dataset} & \multicolumn{2}{c}{DiTing} & \multicolumn{2}{c}{STEAD} \\
            \hline
            Task & Model & MAE & R\textsuperscript{2} & MAE & R\textsuperscript{2} \\
            \hline
            \multirow{3}{*}{\makecell{Back \\ Azimuth}}
            & BAZ network  & 70.83 & 0.255 & 51.78 & 0.517 \\
            & SeisT-L      & \textcolor{blue}{65.98} & \textcolor{blue}{0.302} & \textcolor{blue}{39.76} & \textcolor{blue}{0.647} \\
            & SeisMoLLM    & \textcolor{red}{60.53} & \textcolor{red}{0.357} & \textcolor{red}{26.13} & \textcolor{red}{0.798} \\
            \hline
            \multirow{2}*{Distance} 
            & SeisT-L      & \textcolor{blue}{6.593} & \textcolor{blue}{0.951} & \textcolor{blue}{3.716} & \textcolor{blue}{0.975} \\
            & SeisMoLLM    & \textcolor{red}{6.327} & \textcolor{red}{0.957} & \textcolor{red}{3.456} & \textcolor{red}{0.978} \\
            \hline
            \multirow{3}*{Magnitude}
            & MagNet      & \textcolor{red}{0.351} & \textcolor{red}{0.578} & 0.259 & \textcolor{blue}{0.868} \\
            & SeisT-L     & 0.354 & 0.551 & \textcolor{blue}{0.229} & \textcolor{red}{0.897} \\
            & SeisMoLLM   & \textcolor{blue}{0.352} & \textcolor{blue}{0.552} & \textcolor{red}{0.228} & 0.865 \\
            \hline
            Task & Model & \multicolumn{2}{c}{F1} & \multicolumn{2}{c}{F1} \\
            \hline
            \multirow{4}*{Phase P}
            & PhaseNet          & \multicolumn{2}{c}{62.86} & \multicolumn{2}{c}{97.38} \\
            & EQTransformer     & \multicolumn{2}{c}{63.63} & \multicolumn{2}{c}{97.05} \\
            & SeisT-L           & \multicolumn{2}{c}{\textcolor{blue}{65.33}} & \multicolumn{2}{c}{\textcolor{red}{97.53}} \\
            & SeisMoLLM         & \multicolumn{2}{c}{\textcolor{red}{68.11}} & \multicolumn{2}{c}{\textcolor{blue}{97.47}} \\
            \hline
            \multirow{4}*{Phase S} 
            & PhaseNet          & \multicolumn{2}{c}{46.40} & \multicolumn{2}{c}{\textcolor{blue}{79.27}} \\
            & EQTransformer     & \multicolumn{2}{c}{46.17} & \multicolumn{2}{c}{78.07} \\
            & SeisT-L           & \multicolumn{2}{c}{\textcolor{blue}{48.32}} & \multicolumn{2}{c}{79.14} \\
            & SeisMoLLM         & \multicolumn{2}{c}{\textcolor{red}{51.53}} & \multicolumn{2}{c}{\textcolor{red}{79.60}} \\
            \hline
        \end{tabular}
        \vfill
        \vspace{8pt} 
        \par\noindent\makebox[\linewidth][c]{%
        \parbox{\linewidth}{The F1-scores are written as percentages, and error tolerance in phase P/S picking is 0.1 second. Using seismic waveforms of original length for all the tasks.}
        }
        \end{table}

        Similar to its performance in supervised training, SeisMoLLM demonstrates state-of-the-art (SOTA) performance in the majority of tasks under few-shot settings, achieving optimal results in 12 out of 18 metrics and securing second-best performance in 3 others. Even in cases where it slightly trails the previous best record, SeisMoLLM consistently delivers competitive results. Notably, in challenging tasks related to earthquake location, SeisMoLLM achieves significant improvements. For back-azimuth estimation, it reduces MAE by 8\% and 34\% and enhances R\textsuperscript{2} by 18\% and 23\% in the DiTing and STEAD datasets, respectively. And in epicentral distance estimation, the MAE is reduced by 4\% and 7\%. For phase picking, particularly under the low signal-to-noise ratio of the DiTing dataset, SeisMoLLM improves the F1 score for P-wave and S-wave picking by 3\% and 6\%, respectively. As shown in Figure \ref{few-picking}, it shows robust performance in accurately picking P and S phases, even when handling the waveforms with low signal-to-noise ratios, far epicentral distances, and low magnitude. These results highlight SeisMoLLM's superior generalization capability in few-shot scenarios, especially for complex and noise-sensitive seismic tasks.
        Furthermore, in magnitude estimation, SeisMoLLM performs nearly on par with the best methods, showcasing strong competitiveness and robustness.
    
        Moreover, few-shot setting is more representative of real-world seismic monitoring scenarios compared to full-data or zero-shot settings, as most regions have local data collections can be used for few-shot training. Therefore, few-shot generalization capability is crucial for the practical application of models.
    
    \subsection{Efficiency Analysis}
    
\begin{figure}
        \centering
        \includegraphics[width=0.5\textwidth]{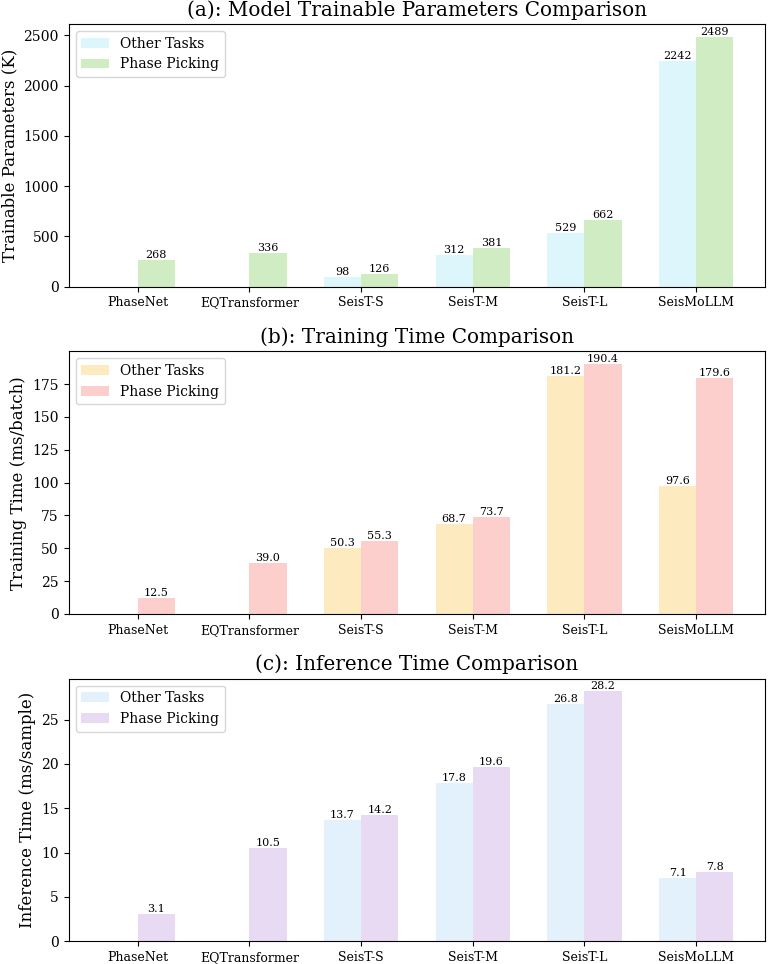}
        \caption{Due to the high similarity in parameter count and efficiency, results for all tasks except phase picking are averaged and categorized as "other tasks," while phase picking that differs from them is classified separately. (a) shows the comparison of trainable model parameters in thousands, (b) compares the training time with a batch size of 128, and (c) compares the inference time for a single waveform sample. The time costs are measured in microseconds.} 
        \label{efficiency}
        \vspace{-0.1cm}
        \end{figure}

        Real-time inference is critical in earthquake monitoring tasks, necessitating a balance between accuracy and time efficiency. With the incorporation of LLM into SeisMoLLM, its efficiency was a focal point of our analysis. We compared SeisMoLLM with baseline small models in terms of the number of trainable parameters and the time costs of training and inference on the DiTing dataset, under identical hardware and code environments. Results summarized in Figure \ref{efficiency}, all time cost values use the average of 500 iterations. Since the model structures and time costs of all tasks except phase picking are very similar, we used their average as the "other" category for comparison.
    
        As shown in Figure \ref{efficiency}.(a), while SeisMoLLM has about 4–20× the parameter count of baseline small models, its training speed, depicted in Figure \ref{efficiency}.(b), surpasses the largest baseline SeisT-L and approaches smaller ones for "other" tasks, demonstrating that despite incorporating LLM, SeisMoLLM still maintains training efficiency comparable to smaller models.
    
        For inference, which is much more important in real-world earthquake monitoring, Figure \ref{efficiency}.(c) reveals that SeisMoLLM is 1.5–3.8× faster than most baselines across all tasks, even reducing the inference time by approximately 50\% compared to the 20× smaller SeisT-S model, achieving an extremely short per-trace inference time of just 7–8 ms. These results confirm that SeisMoLLM not only meets real-time inference requirements but also outperforms baseline small models in inference speed, ensuring practical feasibility for real-world applications.
    
        And we attribute SeisMoLLM's surprisingly high efficiency to three main reasons:
        1. The Convolutional Embedder and Latent Patching compress the input data length by 64×, producing significantly shorter token sequences compared to the intermediate features of other baselines. This substantially alleviates the computational burden on the resource-intensive LLM Blocks. 
        2. Since the backbone LLM Blocks consist of only 3 layers, and the Convolutional Embedder requires just 4 layers thanks to its multi-scale design, SeisMoLLM features a shallower and wider macro-architecture compared to typical deep neural networks, facilitating additional parallel acceleration for many computations inside the model. 
        3. Similar to \cite{ofa}, we believe that the efficient optimizations in HuggingFace's GPT-2 model implementation\cite{huggingface} also helped further reduce the time cost.
    
    \subsection{Ablation Study}
    
        The core idea of our method is to transfer the powerful sequence modeling capability of pre-trained LLM to seismic monitoring in a cross-modal fashion. But recently, there has been debate over whether this paradigm is actually useful for time-series forecasting \cite{llm-effect}. So to evaluate its effectiveness in seismic monitoring, we conducted comprehensive ablation studies following \cite{llm-effect}, which demonstrated the validity of incorporating pre-trained LLM into our method. And the ablations include the following setups:
    
        \textbf{w/o LLM}: Completely remove the LLM Blocks, directly passing the patch embeddings to the final output head. 
        \textbf{LLM2Att}: Replace the LLM Blocks with a randomly initialized multi-head attention layer. 
        \textbf{LLM2Trsf}: Replace the LLM Blocks with a randomly initialized Transformer block. 
        \textbf{w/o pre-train}: Train randomly initialized LLM Blocks from scratch instead of fine-tuning frozen pre-trained parameters.
    
        In addition, we also verified the effectiveness of the Multi-Scale Convolutional Embedder and explored the impact of the number of LLM layers used in SeisMoLLM: 
    
        \textbf{w/o Conv}: Replace the structure preceding the LLM Blocks with a simpler classical design, applying patching with a stride of 32 followed by linear embedding to the input waveforms. 
        \textbf{2 layers}: Use only two GPT-2 layers as the LLM Blocks. 
        \textbf{6 layers}: Add three more layers for a total of six as the LLM Blocks. 
        \textbf{12 layers}: Use all 12 GPT-2 layers as the LLM Blocks.
    
        Due to computational constraints, all these experiments were conducted on the DiTing dataset, and the results are summarized in Table \ref{ablation}.
    
        \begin{table*}\small
        \caption{Results of Ablation Studies on DiTing Dataset \\ \textcolor{red}{red}: worst, \textcolor{blue}{blue}: second worst, \textcolor{Green}{green}:     improvement\label{ablation}}
        \centering
        \setlength{\extrarowheight}{2.5pt}
        \setlength{\tabcolsep}{1.8mm}
        \begin{tabular}{lcccccccccccc}
        \hline
        Task & \multicolumn{2}{c}{Back-azimuth} & \multicolumn{2}{c}{Distance} & \multicolumn{2}{c}{Magnitude} &  \multicolumn{2}{c}{Phase P} & \multicolumn{2}{c}{Phase S} & \multicolumn{2}{c}{Polarity}\\
        \hline
        Model & MAE & $\Delta$ & MAE & $\Delta$ & MAE & $\Delta$ & F1 & $\Delta$ & F1 & $\Delta$ & F1 & $\Delta$ \\
        \hline
        w/o LLM & 41.22 & \textcolor{blue}{+7.67} & 12.64 & \textcolor{red}{+9.66} & 0.267 & \textcolor{red}{+0.103} & 92.77 & \textcolor{blue}{-2.92} & 68.51 & \textcolor{blue}{-4.31} & 93.21 & -1.12 \\
        LLM2Att & 37.48 & +3.93 & 6.124 & \textcolor{blue}{+3.144} & 0.199 & +0.035 & 94.31 & -1.38 & 71.26 & -1.56 & 93.69 & -0.64 \\
        LLM2Trsf & 38.29 & +4.74 & 5.889 & +2.909 & 0.198 & +0.034 & 94.34 & -1.35 & 71.36 & -1.46 & 93.66 & -0.67 \\
        w/o pre-train & 44.50 & \textcolor{red}{+10.95} & 2.909 & \textcolor{Green}{-0.71} & 0.224 & +0.060 & 95.43 & -0.26 & 72.11 & -0.71 & 91.66 & \textcolor{red}{-2.67} \\
        w/o conv & 37.66 & +4.11 & 4.944 & +1.964 & 0.252 & \textcolor{blue}{+0.088} & 91.87 & \textcolor{red}{-3.82} & 68.48 & \textcolor{red}{-4.34} & 92.27 & \textcolor{blue}{-2.06} \\
        \hline
        \textbf{SeisMoLLM} & \textbf{33.55} & \textbf{---} & \textbf{2.980} & \textbf{---} & \textbf{0.164} & \textbf{---} & \textbf{95.69} & \textbf{---} & \textbf{72.82} & \textbf{---} &  \textbf{94.33} & \textbf{---} \\
        \hline
        2 layers & 33.68 & +0.13 & 3.257 & +0.277 & 0.168 & +0.004 & 95.48 & -0.21 & 73.03 & \textcolor{Green}{+0.21} & 94.13 & -0.20 \\
        6 layers & 34.03 & +0.48 & 2.935 & \textcolor{Green}{-0.045} & 0.166 & +0.002 & 95.60 & -0.09 & 72.98 & \textcolor{Green}{+0.16} & 94.26 & -0.07 \\
        12 layers & 33.79 & +0.24 & 2.739 & \textcolor{Green}{-0.241} & 0.167 & +0.003 & 95.48 & -0.21 & 72.79 & -0.03 & 94.32 & -0.01 \\
        \hline
        \end{tabular} \\
        \vspace{5pt}
        The $\Delta$ value shows the performance difference relative to the original SeisMoLLM. Red marks the worst changes, blue the second-worst, and green indicate improvements. The row in bold represents the original performance of SeisMoLLM.
        \end{table*}
    
        Unlike the findings in \cite{llm-effect}, removing the LLM Blocks in SeisMoLLM led to huge performance setback, resulting in either the worst or second-worst results across almost all tasks. This proves the LLM Blocks successfully exerted their powerful feature learning capabilities, confirming their effectiveness. Replacing the LLM Blocks with a multi-head attention layer or Transformer block further reinforced this conclusion. While methods like \cite{ofa}\cite{time-llm} showed improved performance after such replacements, SeisMoLLM consistently outperformed its ablated versions across all tasks. This strongly supports the idea of transferring LLM Blocks to seismic monitoring is effective and impactful.
    
        Compared to baseline models, SeisMoLLM contains significantly more parameters due to the inclusion of LLM Blocks. To verify whether its superior performance solely results from the increased model size, we randomly initialized the LLM Blocks and trained them from scratch instead of fine-tuning frozen pre-trained parameters. This led to noticeable performance drops in five out of six tasks, with severe degradation in challenging tasks such as back-azimuth estimation, and first-motion polarity classification. In these tasks, the setup was even more severe than those caused by structural changes to key components of the model, demonstrating the performance gains of SeisMoLLM stem from the feature learning capabilities of the pre-trained LLM, rather than simple model scaling.
    
        We also explored the impact of the number of LLM layers used in SeisMoLLM, concluding that our choice of stacking 3 layers is appropriate. Increasing the number of layers to 6 or 12 did not effectively improve performance in most tasks and even caused slight declines. This suggests that 3 layers in our model is sufficient for seismic monitoring tasks, and further corroborates simply scaling up the model cannot replicate the substantial performance improvements observed with SeisMoLLM. Conversely, reducing the LLM Blocks to 2 layers resulted in performance degradation across nearly all tasks. In summary, we selected a 3-layer configuration for the LLM Blocks to balance performance and efficiency.

\section{Discussion}

    \subsection{The Idea and Methods of Cross-Modal Transfer}
    
        As deep learning enters the pre-training era, the availability of more unified and high-quality data typically leads to better task performance. This has driven the development of cross-modal transfer as a strategy to overcome performance bottlenecks and generalization challenges in domains with limited pre-training data. Although the seismic domain has accumulated considerable data, the lack of unified datasets suitable for pre-training makes cross-modal transfer a promising and underexplored approach, warranting further investigation by the community.
    
        The challenge in cross-modal transfer lies in maintaining the performance of pre-trained models while bridging the gap between modalities. Here, we discuss two main approaches emerging in the time-series domain:
    
        1. Efficient fine-tuning of pre-trained models: As demonstrated in this work, the powerful capabilities acquired during pre-training are embedded in the model parameters. This approach only uses and fine-tunes only a small part of the pre-trained model to minimize changes in its parameters, thereby mitigating the loss of capacity caused by cross-domain re-training \cite{ofa}\cite{tempo}. Briefly, this method favors adapting the model to the target data modality.
    
        2. Alignment of cross-modal feature: This approach typically utilizes and freezes all pre-trained model parameters to preserve performance, and focuses on aligning feature space between different modalities in the early stages to achieve cross-modal transfer \cite{test}\cite{time-llm}. And in summary, this method favors transforming the target data to align with the pre-trainde model.
    
        We chose the first approach because it achieves excellent results using only parts of the pre-trained model, whereas the second often requires significantly larger and complete pre-trained model to ensure performance \cite{test}\cite{time-llm}, resulting in substantial inefficiencies. Although this fine-tuning strategy complicates training, it simplifies inference and is therefore better suited to the requirements of seismic monitoring.
    
        These two approaches are largely orthogonal and have been successfully combined in recent advanced works \cite{s2ip-llm}\cite{calf}. Building on this foundation, cross-modal transfer holds great potential for application across diverse domains, including seismology.
    
    \subsection{Characteristics of Seismogram and Modality Gap}
    
        In this section, we discuss the characteristics of seismic waveform data and analyze its modality gap in comparison to other common modalities. A salient feature of seismic waveforms is the sparse yet highly concentrated distribution of critical information, with the majority of relevant earthquake event information being densely localized around P- and S-wave arrivals.
    
        In contrast, general time-series data in domains such as traffic, weather, or power systems typically exhibit strong long-term characteristics, including periodicity and trends. These data can often be well represented by combinations of these characteristic components, resulting in smoother and more continuous feature distributions \cite{stl}\cite{tempo}\cite{s2ip-llm}.
    
        From this perspective, seismograms differ substantially from typical time-series data but share similarities with natural language sentences. In both cases, the distribution of effective features, such as keywords that determine semantics, is similarly discrete. This resemblance may partly explain why transferring LLMs to seismic monitoring appears more effective than for other time-series tasks.
    
    \subsection{Limitations and Future Work}
    
        As the first exploration of cross-modal transfer in the seismic monitoring domain, this work has achieved several successes but also faced some inevitable limitations:
    
        1.	Limited exploration of additional modalities: While natural language has been the primary modality explored, other domains, such as speech/audio \cite{whisper}, offer promising pre-trained models for cross-modal transfer to seismology. These models often require pre-processing waveform data into spectrograms for frequency-domain learning that we did not explored in depth. Furthermore, previous work suggests that frequency-domain learning holds significant potential for seismic waveform analysis \cite{seisclip}.
    
        2.	Limited progress in phase picking: As shown in Tables \ref{performance1}, \ref{performance2}, and \ref{few-shot}, SeisMoLLM’s improvements on phase picking are not as substantial as those in other tasks. The phase picking results for the "w/o conv" configuration in Table \ref{ablation} imply that the limitation arises from the inherent nature of the task. Phase picking involves dense predictions for each data point, akin to semantic segmentation or small object detection in computer vision. Convolutional networks excels at these tasks due to their inductive bias that captures local details effectively. In contrast, the backbone LLM Blocks in SeisMoLLM operate at the patch level and cannot fully retain local fine-grained information, resulting in limited improvement on phase picking. This insight may also explain the initial success of PhaseNet \cite{phasenet}.
    
        3.	Limited exploration of multi-task learning: Unifying all seismic monitoring tasks into a single model remains tricky, and this study did not explore this in detail. However, using cross-modal transfer to build unified multi-task models presents a promising approach for future work.
    
        To overcome these limitations, we outline the following future research directions that we hope will inspire further exploration in the community:
        1.	Explore the potential of transferring pre-trained models from other advanced modalities, such as vision\cite{sam} and speech/audio\cite{whisper}, into seismic monitoring. This could be supplemented with the ideas discussed in Discussion A and broaden the applicability of cross-model transfer.
        2.	Utilizing more advanced network backbones, such as using advanced CNNs\cite{convnext} or Swin Transformers\cite{swin-transformer}, which are better suited for both discriminative tasks like earthquake source parameterization and dense prediction tasks like phase picking. These methods have the potential to enable a model to excel at all seismic monitoring tasks.
        3.	Investigate an all-in-one model using methods from multi-task learning, with a focus on adapting cross-modal transfer.
    
\section{Conclusion}

    In this work, we pioneer a novel cross-modal transfer approach to overcome the performance bottleneck and develop a foundation model for seismic monitoring. This approach leverages large-scale pre-training from a large language model, eliminating the need for pre-training from scratch on seismic datasets, which is often hindered by the necessity for vast amounts of data and the heterogeneity across seismic datasets. By fine-tuning a pre-trained LLM and integrating a multi-scale convolutional embedder with latent patching, we introduce SeisMoLLM, effectively transferring the LLM's superior sequence modeling capabilities to seismic monitoring. 
    
    Through extensive experiments on DiTing and STEAD datasets from different regions, SeisMoLLM achieves state-of-the-art performance across multiple tasks, including earthquake location, magnitude estimation, phase picking, and first-motion polarity classification, outperforming existing methods by a significant margin. Moreover, SeisMoLLM demonstrates remarkable few-shot generalization capabilities, enabling strong performance on downstream tasks with minimal fine-tuning data while maintaining efficient training and real-time inference. 
    
    These results highlight the potential of SeisMoLLM as a foundation model for various tasks in seismic monitoring, and reveal the promising prospects of cross-modal transfer and advanced deep learning techniques for progressing seismology research.


{\appendix[Evaluation Metrics]

\setcounter{equation}{0}
\renewcommand{\theequation}{A.\arabic{equation}}


    The evaluation metrics for classification tasks are defined as:
    \begin{equation}
    \label{cls_metrics}
    \begin{split}
    \text{Precision} &= \frac{\text{TP}}{\text{TP} + \text{FP}} \\
    \text{Recall} &= \frac{\text{TP}}{\text{TP} + \text{FN}} \\
    \text{F1 Score} &= 2 \cdot \frac{\text{Precision} \cdot \text{Recall}}{\text{Precision} + \text{Recall}}
    \end{split}
    \end{equation}
    where TP means true positive predictions, FP indicates false positive predictions, and FN is the false negative predictions. 
    
    Precision measures the proportion of correctly identified positive samples among all positives, reflecting the model’s ability to avoid false alarms. Recall evaluates how well the model captures actual positive instances, indicating its ability to minimize missed detections. F1 Score balances precision and recall, making it a useful metric when there is an imbalance between positive and negative samples.

    And for regression tasks, the metrics are as follows:
    \begin{equation}
    \label{reg_metrics}
    \begin{split}
    \text{MAE} &= \frac{1}{n} \sum_{i=1}^{n} \left| y_i - \hat{y}_i \right| \\
    \text{R}^2 &= 1 - \frac{\sum_{i=1}^{n} \left( y_i - \hat{y}_i \right)^2}{\sum_{i=1}^{n} \left( y_i - \bar{y} \right)^2} \\
    \text{MAPE} &= \frac{1}{n} \sum_{i=1}^{n} \left| \frac{y_i - \hat{y}_i}{y_i} \right| \times 100 \\
    \text{RMSE} &= \sqrt{\frac{1}{n} \sum_{i=1}^{n} \left( y_i - \hat{y}_i \right)^2}
    \end{split}
    \end{equation}
    where $\hat{y}_i$ and $y_i$ are predictions and ground truth of the i-th sample, respectively.
    
    MAE (Mean Absolute Error) quantifies the average absolute difference between predictions and actual values, providing an intuitive measure of prediction error. R\textsuperscript{2} (Coefficient of Determination) evaluates the proportion of variance in the target variable explained by the model, indicating its overall goodness of fit. MAPE (Mean Absolute Percentage Error) expresses the prediction error as a percentage, making it useful for evaluating performance across different scales. RMSE (Root Mean Squared Error) penalizes larger errors more than MAE due to the squared term, making it sensitive to outliers while providing a measure of overall error magnitude.

}

\bibliographystyle{IEEEtran}
\bibliography{references}


 




\vfill

\end{document}